\documentclass{article}
\usepackage{spconf}
\usepackage{graphicx}
\usepackage[cmex10]{amsmath}
\usepackage{amsfonts}
\usepackage{amssymb}
\usepackage{dblfloatfix}
\usepackage{xspace}
\usepackage{algorithm,algorithmic}
\usepackage{diagbox}
\usepackage{epstopdf}
\usepackage{color}
\usepackage{pbox}

\newcommand\KM{$k$-means\xspace}

\graphicspath{{./fig/}}

\title{Unsupervised vehicle recognition using incremental reseeding of acoustic signatures}

\twoauthors
{Justin Sunu, Allon G.\ Percus}
{Institute of Mathematical Sciences\\
    Claremont Graduate University\\
    Claremont, CA 91711\\
    justinsunu@gmail.com, allon.percus@cgu.edu}
{Blake Hunter}
{Department of Mathematical Sciences\\
    Claremont McKenna College\\
    Claremont, CA 91711\\
    blake.hunter@claremontmckenna.edu}

\begin{document}
\maketitle

\begin{abstract}
Vehicle recognition and classification have broad applications, ranging from traffic flow management to military target identification.  We demonstrate an unsupervised method for automated identification of moving vehicles from roadside audio sensors. Using a short-time Fourier transform to decompose audio signals, we treat the frequency signature in each time window as an individual data point. We then use a spectral embedding for dimensionality reduction.  Based on the leading eigenvectors, we relate the performance of an incremental reseeding algorithm to that of spectral clustering.  We find that incremental reseeding accurately identifies individual vehicles using their acoustic signatures.
\end{abstract}

\section{Introduction}

Recognizing and distinguishing moving vehicles based on their audio signals are problems of broad interest.  Applications range from traffic analysis and urban planning to military vehicle recognition.  Audio data sets are small compared to video data, and multiple audio sensors can be placed easily and inexpensively.  However,  challenges arise due to equipment as well as to the underlying physics.  Microphone sensitivity can result in disruption from wind and ambient noise.  The Doppler shift can make a vehicle's acoustic signature differ according to its position.


In order to interpret acoustic signatures, one must extract information contained within the raw audio data.  A natural feature extraction method is the short-time Fourier transform (STFT), with time windows chosen large enough that they carry sufficient frequency information but small enough that they can localize vehicle events.  STFT has been used previously for classifying cars vs.\ motorcycles with principle component analysis~\cite{VSSRbFVPCA}, for characterizing $\epsilon$-neighborhoods in vehicle frequency signatures~\cite{VIuWSN}, for vehicle classification based on power spectral density~\cite{AEoFEMfVCBOAS}, and for estimating the fundamental frequency in engine sounds~\cite{SFftCoCVuAS}. Other related feature extraction approaches have included the wavelet transform \cite{AEoFEMfVCBOAS,Wbadomv} and the one-third-octave filter bands \cite{MVNCuMC}.


Our study uses a spectral embedding approach to identify different individual vehicles.  Representing each time window as an individual data point, we define a similarity measure between two points based on the cosine distance between their sets of Fourier coefficients, and then cluster according to the symmetric normalized graph Laplacian~\cite{Luxburg2007}.  We relate the eigenvectors of the Laplacian to a recently proposed clustering method, incremental reseeding (INCRES)~\cite{INCRES}, that iteratively propagates cluster labels across a graph.  We compare the performance of INCRES  with spectral clustering on the vehicle audio data.  We find that both are promising unsupervised methods for vehicle identification, with INCRES correctly clustering 91.7\% of the data points in a sequence of passages of three different vehicles.

\section{Algorithms}

Our clustering algorithms are based on the use of spectral embedding for dimensionality reduction.  
Consider a signal of length $n$, with feature vector $\mathbf{x_i}\in\mathbb{R}^m$ associated with data point $i\in\{1,\dots,n\}$.  A spectral embedding represents data as vertices on a weighted graph, with edge weights $S_{ij}$ expressing a similarity measure between data points $i$ and $j$.  The graph is encoded using the symmetric normalized graph Laplacian matrix~\cite{Luxburg2007}
$$\mathbf{L_s} = \mathbf{I} - \mathbf{D}^{-1/2} \mathbf{S}\mathbf{D}^{-1/2}$$
where $\mathbf{D}$ is a diagonal matrix with $D_{ii}=\sum_j S_{ij}$.

We use this embedding for vehicle identification with two related clustering methods: spectral clustering, and a recently developed incremental reseeding approach~\cite{INCRES}.

\subsection{Spectral clustering}

The eigenvectors of $\mathbf{L_s}$ associated with the leading nontrivial eigenvalues $\lambda_2,\dots\lambda_k$ form a $(k-1)$-dimensional approximation to $\mathbf{x_i}$.  The approximation is justified when the spectral gap $|\lambda_{k+1}-\lambda_k|$ is large, which occurs when the data naturally form $k$ clusters~\cite{Luxburg2007}.  Spectral clustering uses $k$-means to cluster this $\mathbb{R}^{k-1}$ projection of the data.

\subsection{Incremental Reseeding (INCRES) Algorithm}

The INCRES algorithm~\cite{INCRES} is a diffusive method that propagates cluster labels across the graph specified by $\mathbf{L_s}$.  The approach (Algorithm~\ref{Algo1}) is incremental: it plants cluster seeds among nodes, grows clusters from these seeds, and then reseeds among the grown clusters.

\begin{algorithm}[H] \caption{INCRES}\label{Algo1}
  \begin{algorithmic}[1]
    \STATE \textbf{Input} Similarity Matrix $S$, number of clusters $k$, number of iterations $s$
    \STATE \textbf{Initialize} Random partitioning of data
    \FOR {$i=1$ to $s$}
    \STATE PLANT (Initialize random walk matrix)
    \STATE GROW (Propagate the random walk matrix)
    \STATE HARVEST (Finalize the clusters)
    \ENDFOR
  \end{algorithmic}
\end{algorithm}

Since the random walk process is governed by the graph Laplacian, INCRES is closely connected with spectral clustering.  Eigenvectors of $\mathbf{L_s}$ are organized hierarchically: the second eigenvector separates data into two clusters at the coarsest resolution, the third eigenvector identifies a third cluster at a finer resolution, and so on.  An application of INCRES with parameter $k$ propagates $k$ labels through the graph, resulting in $k$ clusters governed by the spectral properties of $\mathbf{L_s}$.

\begin{figure}[htb]
  \centering
  \includegraphics[width=.3\columnwidth]{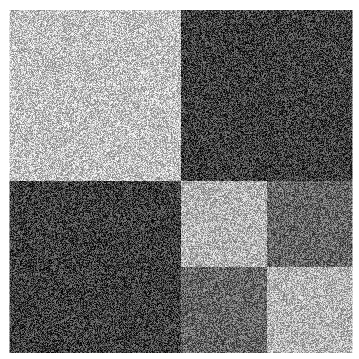}
  \caption{Synthetic similarity matrix with salt and pepper noise.  White represents a similarity value of 1; black represents a similarity value of 0.}
  \label{synth1} 
\end{figure}
\begin{figure}[htb]
  \begin{tabular}{cc}
    \includegraphics[width=.44\columnwidth]{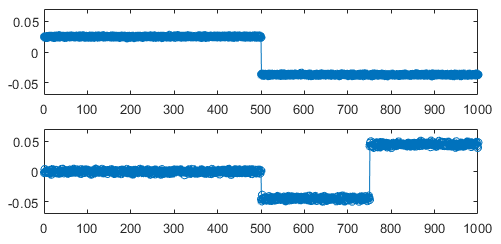} &
    \includegraphics[width=.44\columnwidth]{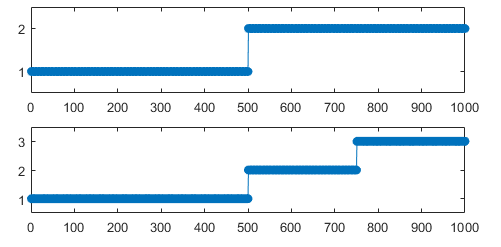}\\
    a) Eigenvectors &
    b) INCRES results
  \end{tabular}
  \caption{2nd and 3rd eigenvectors of $\mathbf{L_s}$ as well as INCRES results, for synthetic similarity matrix.  Note identical separation into two and three clusters.}
  \label{synth2}
\end{figure}

To illustrate the relation between the two algorithms, consider a similarity matrix $S_{ij}$ given by a block matrix form with added ``salt and pepper'' noise.  This is shown in Figure~\ref{synth1}, with lighter colors representing greater similarities.  Figure~\ref{synth2} shows the second and third eigenvectors of $\mathbf{L_s}$, along with the results of INCRES for $k=2$ and $k=3$.  The binary clustering results of both methods split the data into the same larger classes, while the third eigenvector and INCRES at $k=3$ find the same subdivision of one of these classes.  In less straightforward clustering examples, the reseeding process can allow INCRES to learn partitions that are not apparent to spectral clustering.  Furthermore, the formulation of INCRES allows it to be applied even in cases of larger datasets where eigenpairs cannot be easily computed.

\section{Data and Feature Extraction}

Our audio data consists of recordings, provided by the US Navy's Naval Air Systems Command~\cite{AFPC}, of different vehicles moving multiple times through a parking lot at approximately 15mph. The original dataset consists of MP4 videos taken from a roadside camera; we extract the dual channel audio signal, and average the channels together into a single channel.  The audio signal has a sampling rate of 48,000 frames per second.  Video information is used only to ascertain ground truth (vehicle identification) for training data.

\begin{figure}[!b]\centering 

\includegraphics[width=\columnwidth]{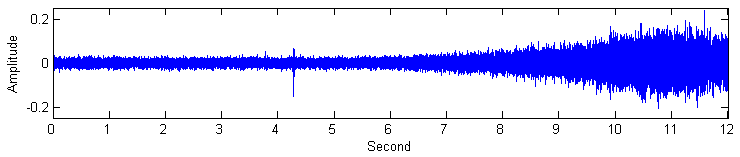}

  \caption{Raw audio signal for vehicle passage.}
  \label{carloc}
\end{figure}

\begin{figure}[!b]\centering 
\begin{tabular}{ccc}
  \multicolumn{3}{c}{\includegraphics[width=\columnwidth]{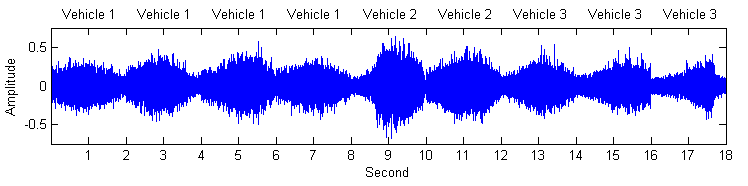}}\\
  \includegraphics[width=.3\columnwidth]{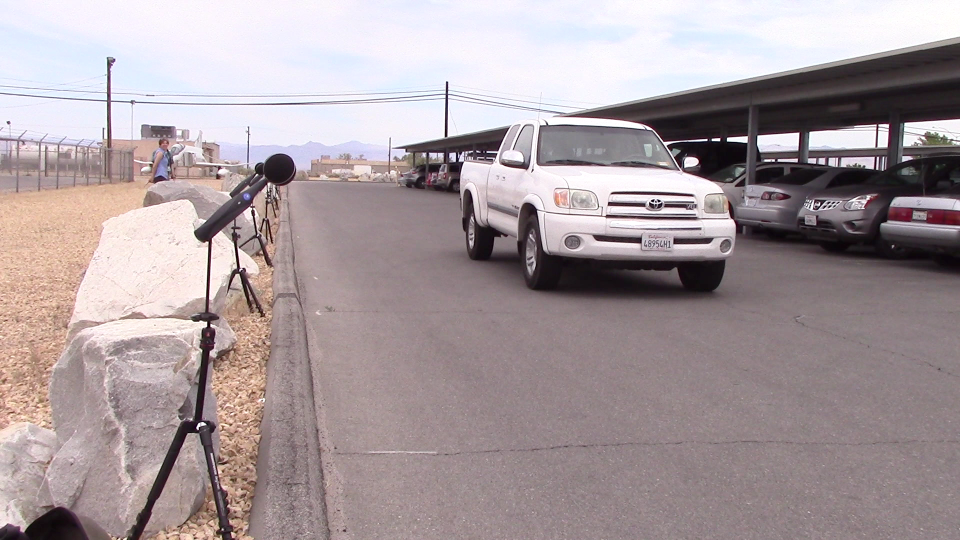} &
  \includegraphics[width=.3\columnwidth]{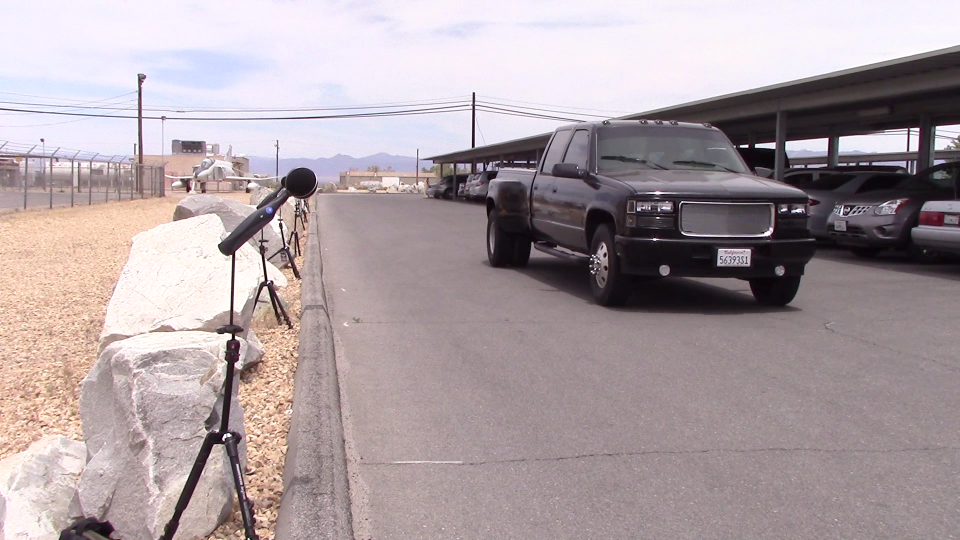} &
  \includegraphics[width=.3\columnwidth]{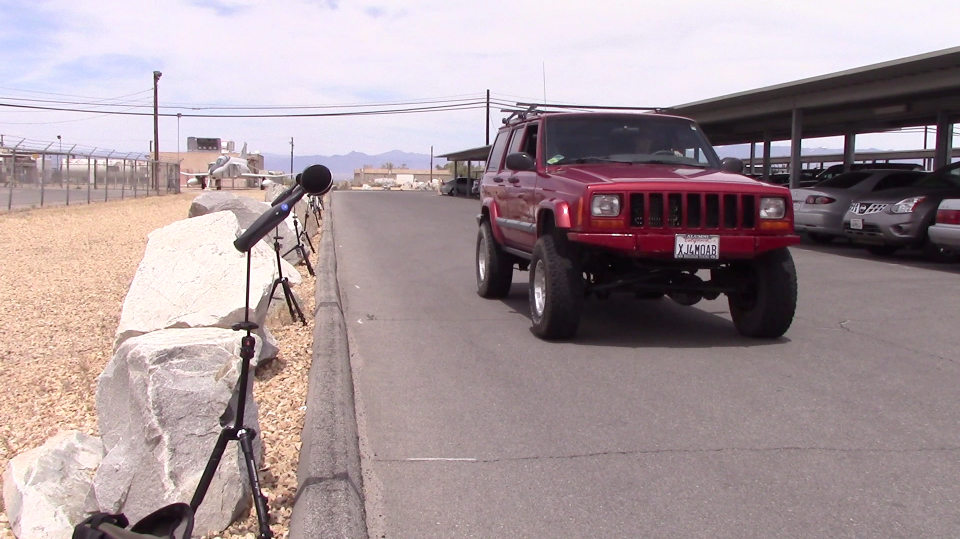} \\
\end{tabular}
  \caption{Raw audio signal for composite data. Images show the three
different vehicles, as seen in accompanying video (not used for
analysis).}
  \label{carloc2}
\end{figure}

\begin{figure*}[!b]
\renewcommand\thefigure{7}
  \begin{tabular}{cc}
    \includegraphics[width=\columnwidth]{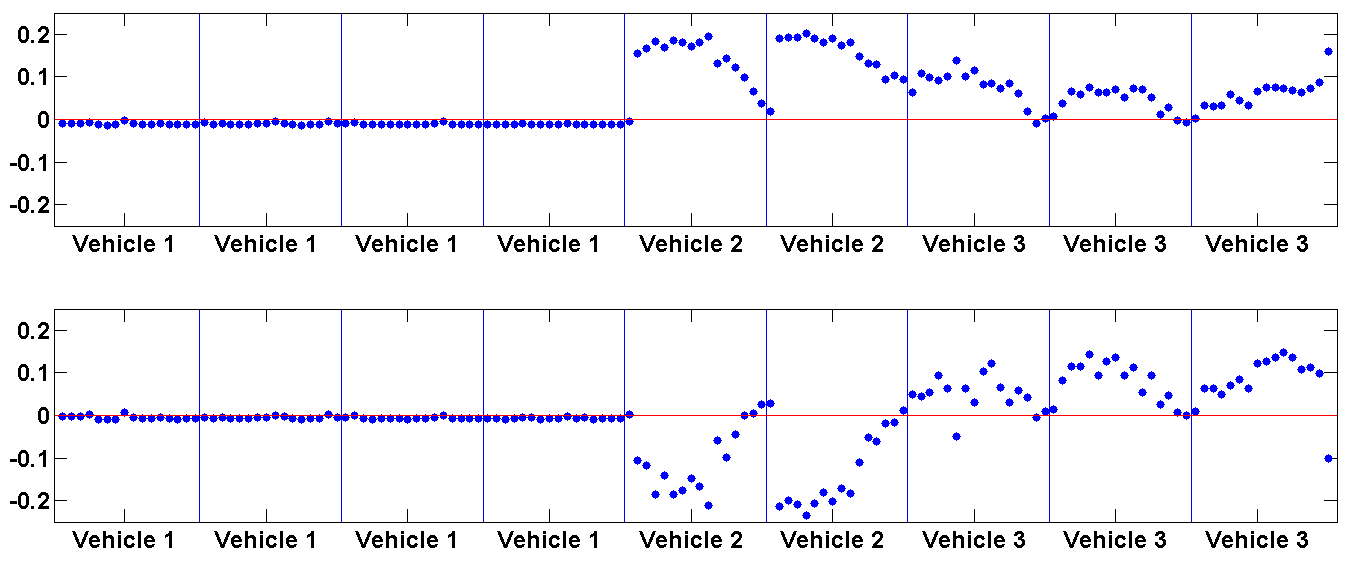} &
    \includegraphics[width=.97\columnwidth]{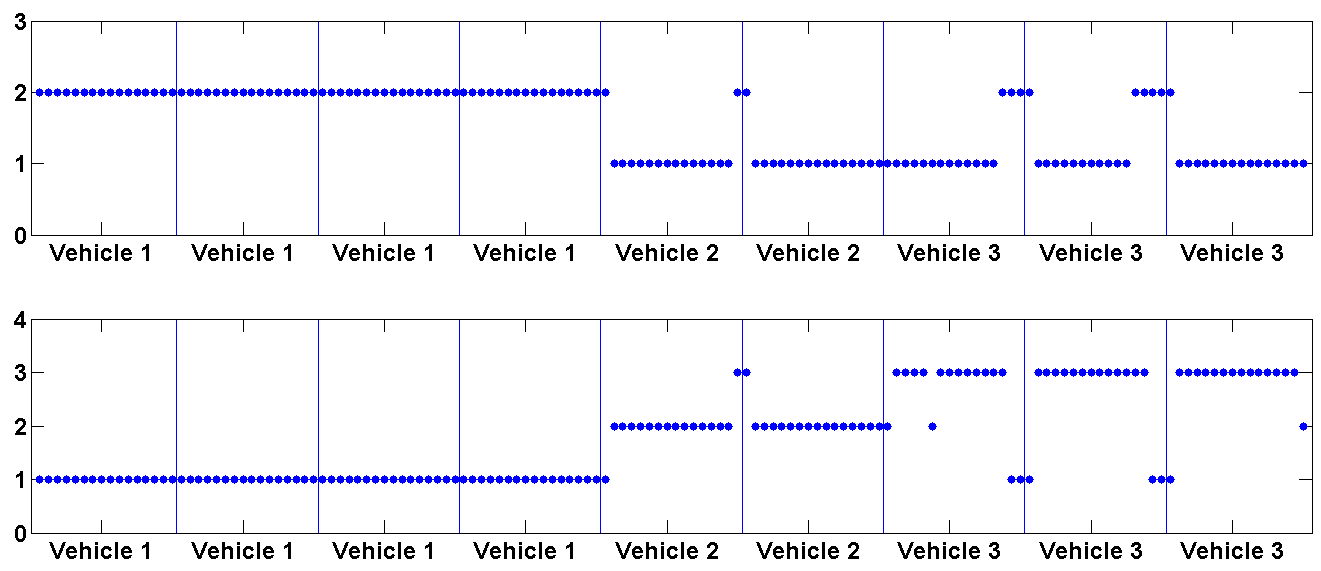}\\
    a) Eigenvectors &
    b) INCRES results
  \end{tabular}
  \caption{2nd and 3rd eigenvectors of $\mathbf{L_s}$ as well as INCRES results, for vehicle data.  }
  \label{fsc39}
\end{figure*}

Each extracted audio signal is a sequence of a vehicle approaching from a distance, becoming audible after 5 or 6 seconds, passing the microphone after 10 seconds, and then leaving.  An example of the raw audio signal is shown in Figure~\ref{carloc}.  We form a composite sequence, shown in Figure~\ref{carloc2}, from multiple passages of three different vehicles (a white truck, black truck, and jeep), cropping the two seconds where the vehicle is closest to the camera.  The goal is to test the clustering algorithm's ability to differentiate the vehicles.

We preprocess the data by grouping audio frames into larger windows.  With windows of $1/8$ of a second, or 6000 frames, we find both a sufficient number of windows and sufficient information per window.  While there is no clear standard in the literature, this window size is comparable to those used in other studies~\cite{VSSRbFVPCA}.  Discontinuities between successive windows can in some cases be reduced by applying a weighted window filter such as a Hamming filter, or by allowing overlap between windows~\cite{VSSRbFVPCA}.  However, in our study we found no conclusive benefit from either of these, and therefore used standard box windows with no overlap.



\begin{figure}[t]
\renewcommand\thefigure{5}
  \centering
  \includegraphics[width=\columnwidth]{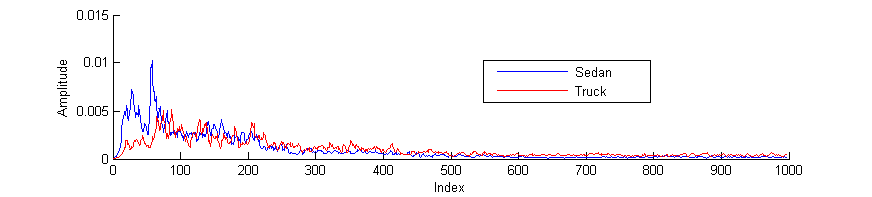}
  \caption{First 1000 Fourier coefficients for a car and a truck, after applying a moving mean of size 5.}
  \label{f16_3} 
\end{figure}

Relevant features are extracted from the raw audio signal using the short-time Fourier transform (STFT). 
The Fourier decomposition contains 6000 symmetric coefficients, leaving 3000 usable coefficients.  Figure \ref{f16_3} shows the first 1000 Fourier coefficients for a time window representing a sedan passing, and a time window representing a truck passing, both in similar positions.  Note that a clear frequency signature is apparent for each vehicle, with much of the signal concentrated within the first 200 coefficients.

Each time window in the audio signal is taken as an independent data point to be clustered: we define the feature vector $\mathbf{x_i}\in\mathbb{R}^m$ as the set of $m$ Fourier coefficients associated with that window.  Since many of these coefficients are relatively insignificant, we consider the cosine distance measure between data points
$$d_{ij} = 1-\frac{\mathbf{x_i}\cdot
\mathbf{x_j}}{\|\mathbf{x_i}\| ~ \|\mathbf{x_j}\|}.$$
We then construct an $M$-nearest neighbor graph, where the edge $\{i,j\}$ is present if $j$ is among the $M$ closest neighbors of $i$ or vice-versa, for a fixed value of $M$. Following standard methods~\cite{Luxburg2007}, the similarity $S_{ij}$ is taken to be a Gaussian function of distance,
$$S_{ij}= e^{-d_{ij}^2/\sigma_i^2},$$
where $\sigma_i$ is defined adaptively~\cite{ZMP} as the distance to vertex $i$'s $M$th neighbor.

\section{Results}

Our composite vehicle dataset contained 18 seconds of raw audio, resulting in $n=144$ data points each representing 1/8-second time windows.  We used only the first $m=1500$ Fourier coefficients.  We set $M=15$ for the $M$-nearest neighbor graph, so that neighborhoods contain the 16 data points used in the 2-second clips of a single vehicle passage.

\begin{figure}[t]
\renewcommand\thefigure{6}
  \centering
  \includegraphics[width=.9\columnwidth]{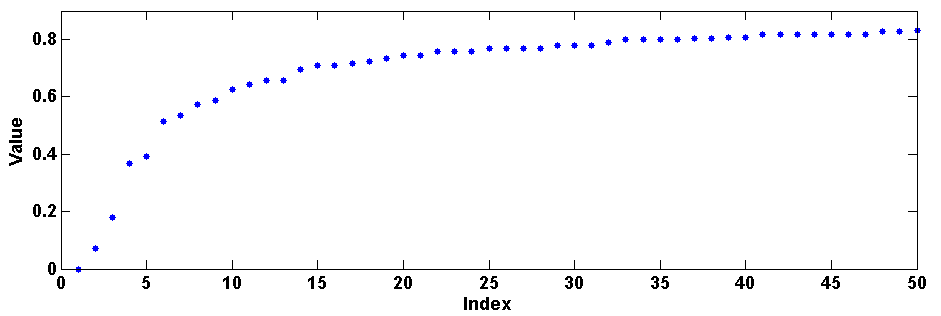}
  \caption{Spectrum of $\mathbf{L_s}$ for vehicle data. Largest gap is after third eigenvalue.}
  \label{fsc31} 
\end{figure}

Figure~\ref{fsc31} shows the eigenvalues of the Laplacian for the vehicle data.  The largest gap follows the third eigenvalue, consistent with three clusters representing the three vehicles actually present in the data.  We therefore set $k=3$ for both spectral clustering and INCRES.

Figure \ref{fsc39} shows the second and third eigenvectors of $\mathbf{L_s}$, along with typical results of INCRES for $k=2$ and $k=3$ (INCRES is stochastic, but results vary little from run to run).  As in our earlier synthetic example, the second eigenvector and $k=2$ INCRES result provide comparable binary separations of the data.  Thresholding the eigenvector just above zero would place all of the vehicle 1 data in one cluster, and most of the vehicle 2 and 3 data in the other cluster (the exceptions are primarily data points at the beginning and end of a vehicle passage, where the signal is weakest).  The third eigenvector mostly distinguishes vehicle 2 (negative values) and vehicle 3 (positive values).  The $k=3$ INCRES result recognizes the three vehicles very accurately, and is discussed below.

\begin{figure}[t]
  \centering
  \includegraphics[width=.9\columnwidth]{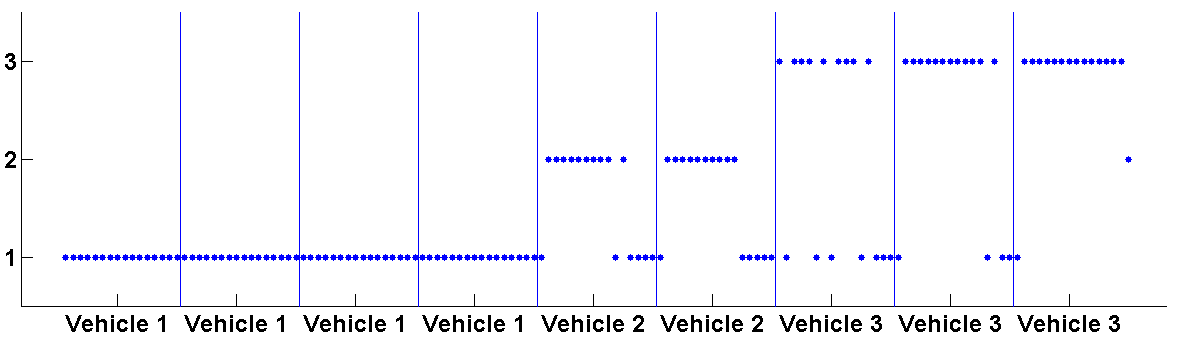}
  \caption{\KM on third eigenvector of $\mathbf{L_s}$ for vehicle data.}
  \label{fsc32} 
\end{figure}

Note that unlike in the straightforward  synthetic data problem, the third eigenvector is not by itself sufficient to separate the three clusters.  Figure~\ref{fsc32} shows the results of \KM clustering, with $k=3$, on the third eigenvector alone.  While all vehicle 1 data points are clustered together, a significant fraction of vehicle 2 and 3 data points are incorrectly placed in that cluster as well.

\begin{figure}[t]
  \centering
  \includegraphics[width=.9\columnwidth]{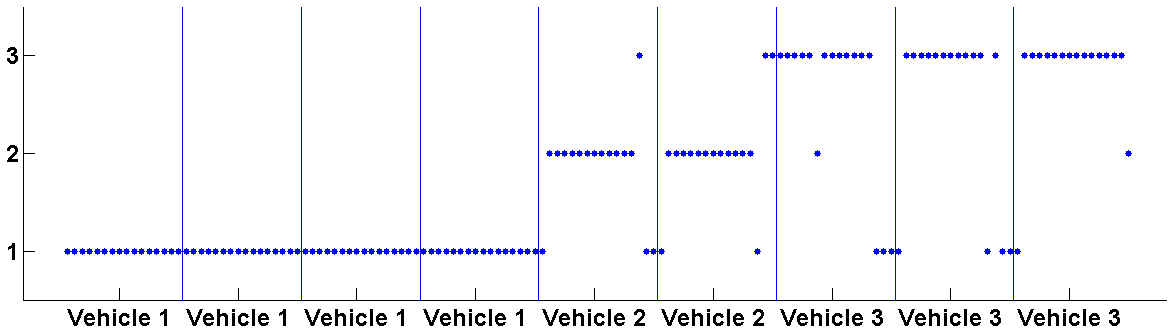}
  \caption{\KM on second and third eigenvectors of $\mathbf{L_s}$ (standard spectral clustering) for vehicle data.}
  \label{fsc33} 
\end{figure}

Figure~\ref{fsc33} shows results of the more conventional spectral clustering method, using \KM on the $\mathbb{R}^2$ projection of the data given by the 2nd and 3rd eigenvectors.  The inclusion of the 2nd eigenvector is sufficient to cluster the vast majority of vehicle 2 and 3 data points correctly.

Tables~\ref{tab2} and \ref{tab3} interpret the spectral clustering results of Figure~\ref{fsc33} and the INCRES $k=3$ results of Figure~\ref{fsc39}b as classifications.  Both methods classify all of vehicle 1 correctly. but INCRES performs noticeably better than spectral clustering on vehicle 2, and they perform comparably on vehicle 3.  Overall purity scores are 87.5\% for spectral clustering, and 91.7\% for INCRES, with misclassifications again occurring primarily at the beginning or end of a vehicle passage.






\begin{table}[t]
  \caption{Vehicle clustering results using spectral clustering.}
  \label{tab2} \centering
  \small
  \begin{tabular}{|c|c|c|c|}
    \hline
\backslashbox{True}{Obtained\\cluster} & \pbox{2in}{Vehicle 1 \\ \centering (w. truck)} & \pbox{2in}{Vehicle 2 \\ \centering (b. truck)} & \pbox{2in}{Vehicle 3 \\ \centering (jeep)} \\
\hline
    Vehicle 1 (white truck)   &     64      &      0      &  0   \\ \hline
    Vehicle 2 (black truck)   &      5      &     24      &  3   \\ \hline
    Vehicle 3 (jeep)       &      8      &      2      &  38  \\ \hline
  \end{tabular}
\end{table}


\begin{table}[t]
  \caption{Vehicle clustering results using INCRES with $k=3$.}
  \label{tab3} \centering
  \small
  \begin{tabular}{|c|c|c|c|}
    \hline
    \backslashbox{True}{Obtained\\cluster} & \pbox{2in}{Vehicle 1 \\ \centering (w. truck)} & \pbox{2in}{Vehicle 2 \\ \centering (b. truck)} & \pbox{2in}{Vehicle 3
\\ \centering (jeep)} \\
\hline
    Vehicle 1 (white truck)   &     64      &      0      &  0   \\ \hline
    Vehicle 2 (black truck)   &      1      &     29      &  2   \\ \hline
    Vehicle 3 (jeep)       &      6      &      3      &  39  \\ \hline
  \end{tabular}
\end{table}

\section{Conclusions}

We have presented a method to identify moving vehicles from audio recordings, by clustering their frequency signatures with an incremental reseeding method (INCRES)~\cite{INCRES}.  We decompose the audio signal with a short-time Fourier transform (STFT), and treat each $1/8$-second time window as an individual data point.  We then apply a spectral embedding and consider the symmetric normalized graph Laplacian.  We find that spectral clustering, which uses the leading eigenvectors of the Laplacian, correctly clusters 87.5\% of the data points. INCRES, which directly uses the Laplacian to construct a random walk on the graph, correctly clusters 91.7\% of the data points.  Almost all incorrectly clustered points lie at the very beginning or very end of a vehicle passage, when the vehicle is furthest from the recording device.  The vast majority of data points result in correct vehicle recognition.

We observe that there is a close relation between the $k$th eigenvector and the INCRES output for $k$ clusters.  This suggests that clustering results might be improved by simultaneously taking the INCRES output for 2 through $k$ clusters, and then using \KM on this $\mathbb{R}^{k-1}$ projection of the data just as spectral clustering does on the 2nd through $k$th eigenvectors.  While doing so does not noticeably change our INCRES $k=3$ results, the difference could be significant for larger values of $k$.  This could be tested, using a dataset with a larger number of vehicles.

Finally, we note that, since time windows are treated as independent data points, our approach ignores most temporal information.  Explicitly taking advantage of the time-series nature of our data in the clustering algorithm could improve results, by clustering data points according not only to their own frequency signatures but also to those of preceding or subsequent time windows.  Furthermore, while the STFT is a standard method for processing audio signals, it suffers from two drawbacks: the use of time windows imposes a specific time scale for resolving the signal that may not always be the appropriate one, and vehicle sounds may contain too many distinct frequencies for the Fourier decomposition to yield easily learned signatures.  These difficulties may best be addressed by using multiscale techniques such as wavelet decompositions that have been proposed for vehicle detection and classification~\cite{AEoFEMfVCBOAS,Wbadomv}, as well as more recently developed sparse decomposition methods that learn a set of basis functions from the data~\cite{SST,HouShi,EWT,ChuiMhaskar}.

\bibliographystyle{IEEEbib}
\bibliography{biblio}

\end{document}